\documentclass[10pt,twocolumn,letterpaper]{article}

\usepackage[pagenumbers]{cvpr}              % To produce the CAMERA-READY version
\usepackage{float}
\usepackage{amsmath}
\usepackage{booktabs}
\usepackage{pifont}
\renewcommand{\and}{\hspace{0.6cm}}

\usepackage{subfiles}

\definecolor{cvprblue}{rgb}{0.21,0.49,0.74}
\usepackage[pagebackref,breaklinks,colorlinks,citecolor=cvprblue]{hyperref}
\usepackage{multirow}

\def\paperID{16917} % *** Enter the Paper ID here
\def\confName{CVPR}
\def\confYear{2026}

\title{Dynamic Visual SLAM using a General 3D Prior}

\author{
Xingguang Zhong$^1$
\and
Liren Jin$^1$
\and
Marija Popovi{\'c}$^2$
\and
Jens Behley$^1$
\and
Cyrill Stachniss$^{1,3}$\\
{\small $^1$Center for Robotics, University of Bonn, Germany $^2$MAVLab, TU Delft, the Netherlands}\\
{\small $^3$Lamarr Institute for Machine Learning and Artificial Intelligence, Germany}}\vspace{-0.1cm}
\begin{document}

\maketitle

\begin{abstract}
  Reliable incremental estimation of camera poses and 3D reconstruction is key to enable various applications including robotics, interactive visualization, and augmented reality.
  However, this task is particularly challenging in dynamic natural environments, where scene dynamics can severely deteriorate camera pose estimation accuracy.
  In this work, we propose a novel monocular visual SLAM system that can robustly estimate camera poses in dynamic scenes.
  To this end, we leverage the complementary strengths of geometric patch-based online bundle adjustment and recent feed-forward reconstruction models.
  Specifically, we propose a feed-forward reconstruction model to precisely filter out dynamic regions, while also utilizing its depth prediction to enhance the robustness of the patch-based visual SLAM.
  By aligning depth prediction with estimated patches from bundle adjustment, we robustly handle the inherent scale ambiguities of the batch-wise application of the feed-forward reconstruction model.
  Extensive experiments on multiple tasks show the superior performance of our proposed method compared to state-of-the-art approaches.
\end{abstract}

%%%%%%%%%%%%%%%%%%%%%%%%%%%%%%%%%%%%%%%%%%%%%%%%%%%%%%%%%%%%%%%%
\section{Introduction}

\begin{figure}[!t]
    \centering
    \includegraphics[width=\linewidth]{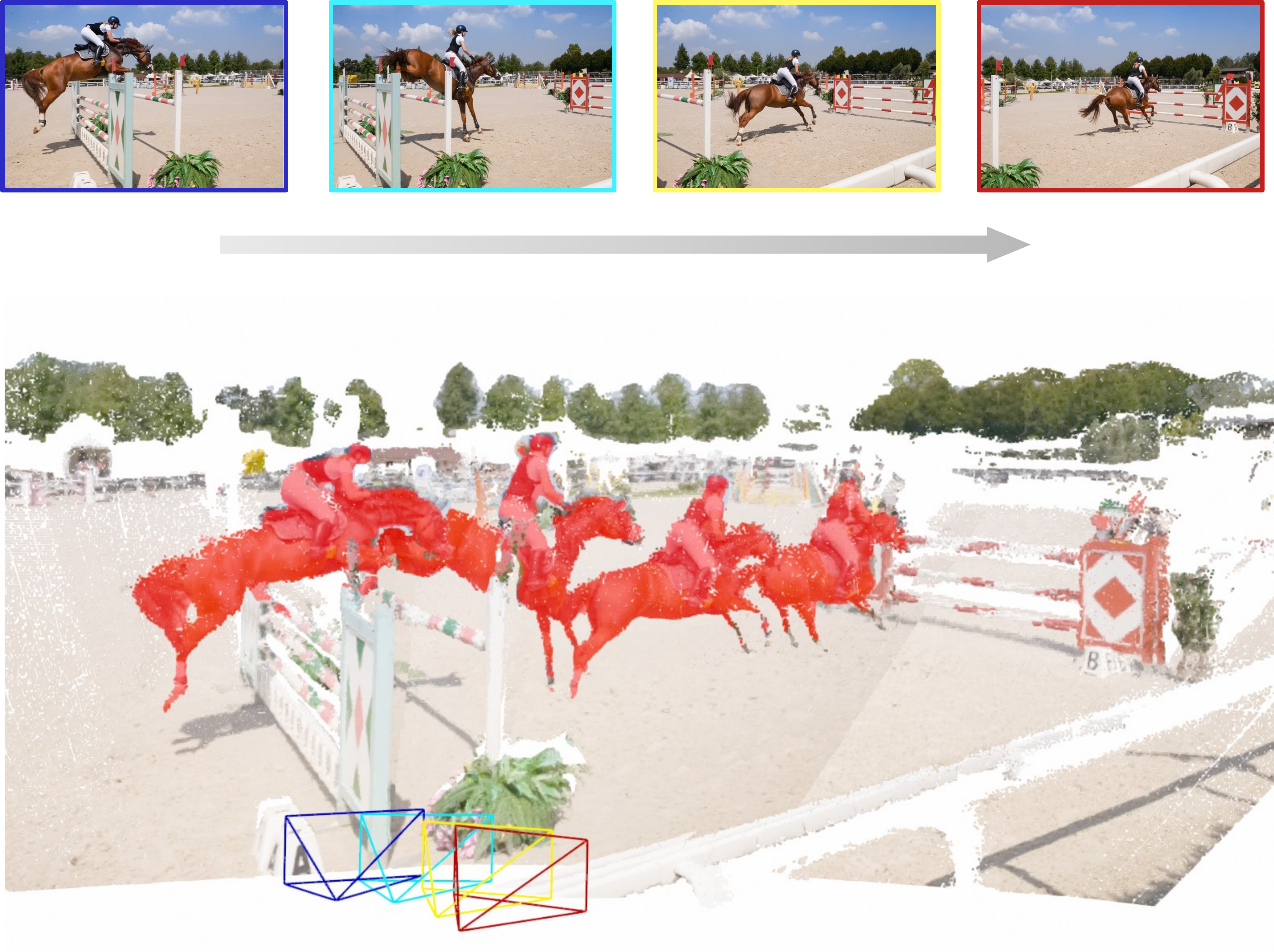}
    \caption{Given an input image sequence, our SLAM system tightly integrating a conventional geometric pipeline with a feed-forward reconstruction model can incrementally perform moving object segmentation (red masks), scale-consistent depth estimation, and camera pose estimation.} \label{motivation}
    \vspace{-0.5cm}
\end{figure}

%%%%%%%%%%%%%%%%%%%
%% WHY: 
% First, answer the WHY question: Why is that relevant? Why should I be
% motivated to read the paper? Why should I care? (1 paragraph, 2-5 sentences)

% A fundamental task in computer vision is estimating the pose of a camera from a sequence of images, which involves recovering the scene geometry as a map and localizing the camera in this map.
% This so-called simultaneous localization and mapping~(SLAM) is required in various application areas, including robotics, computer graphics, enabling ...

Robust estimation of camera poses from a sequence of RGB images is a fundamental problem in computer vision and robotics. 
In particular, incremental pose estimation using only the available camera frames is essential for providing real-time pose information to downstream applications such as safe robot navigation, interactive visualization, and immersive augmented reality. 
Over the past decades, numerous visual odometry~\cite{teed2023nips,forster2014icra,yang2020cvpr,stumberg2018icra} and simultaneous localization and mapping~(SLAM) methods~\cite{campos2021tro,cieslewski2018icra,shin2018icra,vidal2018ral,concha2016icra,mur-artal2017tro} have been proposed, with recent approaches increasingly incorporating learning-based components and scene representations to enhance efficiency, robustness, and generalization~\cite{teed2021neurips,yang2018eccv,lipson2025eccv}.

Despite their effectiveness, most visual odometry and SLAM approaches assume a static environment, an assumption that rarely holds in real-world settings, particularly in dynamic urban scenes. 
The presence of moving objects often leads to incorrect data associations, resulting in inaccurate camera pose estimates and degraded map quality.
In addition, these methods typically perform online sparse reconstruction for computational efficiency, which limits their ability to capture fine-grained scene geometry. 
At the same time, the absence of geometric priors further constrains their robustness and stability under challenging conditions in a monocular setup.

Recently, feed-forward reconstruction models~\cite{wang2025cvpr-vvgg} have been proposed to directly regress dense scene structure and camera poses from sequences of RGB images, thereby eliminating the need for modular system design. 
Trained on large-scale multi-view datasets, these models learn rich geometric priors, achieving impressive performance in challenging scenarios where traditional visual odometry or SLAM systems often fail. 
Moreover, this learning-based paradigm offers extensibility, allowing the integration of additional prior knowledge such as semantics~\cite{zust2025iccv} or moving object segmentation~\cite{zhang2025iclr,chen2025iccv}.
Motivated by this, we pose the following question: Can the priors encoded in feed-forward reconstruction models be effectively exploited to overcome the limitations commonly faced by visual odometry and SLAM systems?

In this work, we propose a visual SLAM framework that tightly integrates a conventional geometric pipeline with feed-forward reconstruction models.
We employ patch-based bundle adjustment for accurate camera pose estimation, while leveraging a feed-forward reconstruction model to detect and exclude moving objects via its learned geometric and semantic priors.
This integration goes beyond simple combination and exploits the complementary strengths of both paradigms.
First, a keyframe-based scale estimation, enabled by patch-based bundle adjustment, resolves the scale drift caused by incremental usage of feed-forward reconstruction models.
Second, depth and confidence outputs from the feed-forward reconstruction model further robustify the patch-based bundle adjustment.
Together, these components enable reliable handling of scene dynamics and accurate camera tracking, leading to strong performance across multiple downstream tasks, as demonstrated in Fig.~\ref{motivation} and our extensive experimental evaluation.

% Link to figure somewhere
% See \figref{fig:motivation} for an example.
In summary, the main contributions of this paper are:
\begin{itemize}
%   \item We build upon and modify an existing feed-forward reconstruction model to achieve accurate multi-frame moving object segmentation with limited data and training computational resources.
  \item We propose a feed-forward reconstruction model with integrated moving object segmentation, enabling our SLAM system to robustly handle dynamic scenes.
  \item We propose a unified framework that integrates depth prior from the feed-forward reconstruction model with geometric SLAM through an uncertainty-aware bundle adjustment mechanism, enabling superior pose estimation performance in dynamic environments.
  \item We address scale ambiguities inherent from incremental usage of the feed-forward reconstruction model through scale alignment with patch-based SLAM, achieving scale-consistent depth estimation over long sequences.
\end{itemize}
We release the code~\footnote{Code: \url{https://github.com/PRBonn/Pi3MOS-SLAM}} to improve reproducibility.

% \begin{itemize}
%   \item We combine a moving object mask prior of a feed-forward 3D reconstruction model into an optical flow-based or patch-based visual SLAM system to improve its performance in dynamic environments.
%   \item The geometric slam method can also help the feed-forward model to work on longer sequences, keep a consistent depth prediction, with limited GPU memory. 
%   \item Our extensive experimental evaluation shows superior performance of our approach for multiple downstream tasks in terms of quality and accuracy: (1) moving object segmentation, (2) monocular pose estimation and reconstruction of the static environment, and (3) depth estimation of a video sequence.
% \end{itemize}

%%%%%%%%%%%%%%%%%%%%%%%%%%%%%%%%%%%%%%%%%%%%%%%%%%%%%%%%%%%%%%%%
\section{Related Work}
\label{sec:related_work}
\textbf{Visual SLAM in Dynamic Environments.}
Visual odometry~\cite{teed2023nips,forster2014icra,yang2020cvpr,stumberg2018icra,engel2018pami, tartanvo2020corl} and visual SLAM~\cite{campos2021tro,cieslewski2018icra,shin2018icra,vidal2018ral,concha2016icra,mur-artal2017tro,campos2021tro,engel2014eccv, teed2021neurips} are classical topics in computer vision and robotics with many advances over the last decades.
However, most visual SLAM systems do not explicitly handle moving objects and instead treat them as outliers, which makes it difficult for these methods to remain stable in highly dynamic environments.
To address this limitation, several works~\cite{bescos2018ral,li2025ral-dsrt,ruenz2018ismar} integrate object detectors or segmentation networks to distinguish common dynamic objects such as humans from the static background. However, such approaches struggle to generalize to unseen categories of moving objects and they cannot differentiate motion states within the same object category.
%
%Some methods~\cite{palazzolo2019iros,dai2022tpami} suppress dynamic objects using geometric cues. These approaches assume moving objects occupy only a small part of the image and require depth measurements.
%
More recent approaches~\cite{chen2024cvpr-lvlt,chen2025iccv-botb,xiao2025iccv} distinguish static and dynamic tracks by analyzing point trajectories over multiple frames. But they rely on the offline segment anything model (SAM)~\cite{kirillov2023iccv, ravi2024iclr} to generate object masks, preventing them from providing an online static map for downstream tasks.
WildGS-SLAM~\cite{zheng2025cvpr-wsmg} builds on NeRF on-the-go~\cite{ren2024cvpr-notg} and WildGaussians~\cite{kulhanek2024neurips}, using \mbox{DINOv2~\cite{oquab2024tmlr}} features and differentiable rendering to train an MLP online for dynamic object segmentation. It requires multiple observations to initialize the MLP reliably, which limits performance under fast sensor motion.
A method closely related to ours, MegaSaM~\cite{li2025cvpr-mafa}, trains a neural network on top of DROID-SLAM~\cite{teed2021neurips} to predict motion probabilities. Its segmentation is limited by network capacity, and as an offline method with high memory usage, it is difficult to apply in online robotic settings.

\textbf{Feed-Forward Reconstruction Models.} 
Given a set of input images, feed-forward reconstruction models can directly infer scene structure and camera poses in a single forward pass, opening new possibilities for 3D reconstruction.
Starting from DUSt3R~\cite{wang2024cvpr-dg3v} and MASt3R~\cite{leroy2024eccv}, which are limited to two-frame inputs, feed-forward reconstruction models have recently made significant progress.
This has led to the development of offline multi-view models~\cite{yang2025cvpr, wang2025cvpr-vvgg, wang2025arxiv-pspe} and incremental models~\cite{wang2024threedv, liu2025cvpr-srtd, wang2025cvpr-c3pm, Wu2025arxiv-ps3r, zhuo2025arxiv-s4gt}.

In an offline multi-view setup, VGGT~\cite{wang2025cvpr-vvgg} and $\pi^3$~\cite{wang2025arxiv-pspe} leverage 2D foundation model \mbox{DINOv2~\cite{oquab2024tmlr}} and self-attention mechanisms to demonstrate state-of-the-art 3D reconstruction performance, even in challenging scenarios with low texture and limited overlap where traditional SLAM or structure-from-motion methods struggle. 
However, a key limitation is that both memory consumption and inference latency grow rapidly with increasing number of input views, which restricts their direct application in online settings. Although some methods~\cite{maggio2025arxiv, deng2025arxiv} adopt mechanisms to perform local map prediction through multi-view feed-forward inference followed by submap fusion, they remain susceptible to accumulated drift, particularly in the presence of moving objects.
Incremental methods, such as CUT3R~\cite{wang2025cvpr-c3pm}, encode local scene information into spatial memory, enabling per-frame predictions conditioned on the evolving memory state. This design allows constant time and memory consumption during inference. 
However, on long sequences, these methods still inevitably exhibit significant pose estimation errors and trajectory drift, falling short of the stability achieved by classical visual SLAM.

Another encouraging advantage of feed-forward models is their robustness in dynamic environments. Through training on large-scale dynamic scene datasets, models such as MonST3R~\cite{zhang2025iclr}, CUT3R~\cite{wang2025cvpr-c3pm}, and $\pi^3$~\cite{wang2025arxiv-pspe} can stably predict camera poses and depth even in highly dynamic scenarios. This observation motivates us to leverage the prior information extracted by these models to address dynamic visual SLAM challenges.

% By combining the strengths of feed-forward reconstruction models and geometric SLAM, our approach enables robust online operation in dynamic environments.

%%%%%%%%%%%%%%%%%%%%%%%%%%%%%%%%%%%%%%%%%%%%%%%%%%%%%%%%%%%%%%%%
\section{Our Approach}
\label{sec:Method}

Given an image sequence $\mathcal{I} = \{\mathbf{I}_i \in \mathbb{R}^{H\times W\times 3}\}_{i=1}^{N}$ captured in a dynamic environment, 
the goal of this work is to robustly estimate camera poses $\mathbf{T}_i \in SE(3)$ and scale-consistent dense depth maps $\mathbf{D}_i \in \mathbb{R}^{H \times W}$ online.
We build our SLAM system upon a monocular SLAM system DPV-SLAM~\cite{teed2023nips, lipson2024eccv} and our proposed feed-forward reconstruction model $\pi^3_{\text{mos}}$.
Specifically, $\pi^3_{\text{mos}}$ can accurately segment moving objects and predict a high-quality depth map at the same time. We integrate these priors within an optical-flow based monocular SLAM framework to achieve more accurate and stable camera pose estimation over long sequences in dynamic scenes.

In the following sections, we first briefly summarize the core architecture of DPV-SLAM, then introduce $\pi^3_{\text{mos}}$ and how it achieves moving object segmentation. Finally, we present our fusion strategy and key improvements that combine the strengths of both systems.

\subsection{Patch-Based Visual SLAM formulation}
\label{subsec:dpvo}
%
% Compared with its predecessor DROID-SLAM~\cite{teed2021neurips}, which establishes inter-frame constraints using dense optical flow, DPVO tracks a fixed number of sparse patches per frame, significantly reducing both memory and computational cost. 
% %
% More specifically, 
%
Given two images $\{\mathbf{I}_i, \mathbf{I}_j\}$, DPV-SLAM samples $K$ patches $P_{1...K}^i \in \mathbb{R}^2$ from image $\mathbf{I}_i$, 
and then uses a lightweight neural network to track their corresponding pixels $P_{1...K}^{ij} \in \mathbb{R}^2$ in image $\mathbf{I}_j$. 
After establishing these correspondences, we perform sliding-window bundle adjustment (BA), 
where the optimization objective is the reprojection error, defined as:
\begin{equation}
 \mathcal{L}_{\text{BA}} = \sum_{(i,j)\in \mathcal{E}} \sum_{k=1}^{K} 
\left\|
P_k^{ij} - \pi\!\left(\mathbf{T}_j \mathbf{T}_i^{-1} \pi^{-1}\!(P_k^i, d_k^i) \right)
\right\|^2,
\label{eq:BA}
\end{equation}
where $\mathcal{E}$ denotes the set of frame pairs within the window, $\mathbf{T}_i \in SE(3)$ is the camera pose of frame $i$,
$d_k^i$ is the inverse depth of patch $P_k^i$ in frame $i$, $\pi(\cdot)$ and $\pi^{-1}(\cdot)$ represent the camera projection and back-projection functions.
\begin{figure}[!t]
    \centering
    \includegraphics[width=\columnwidth]{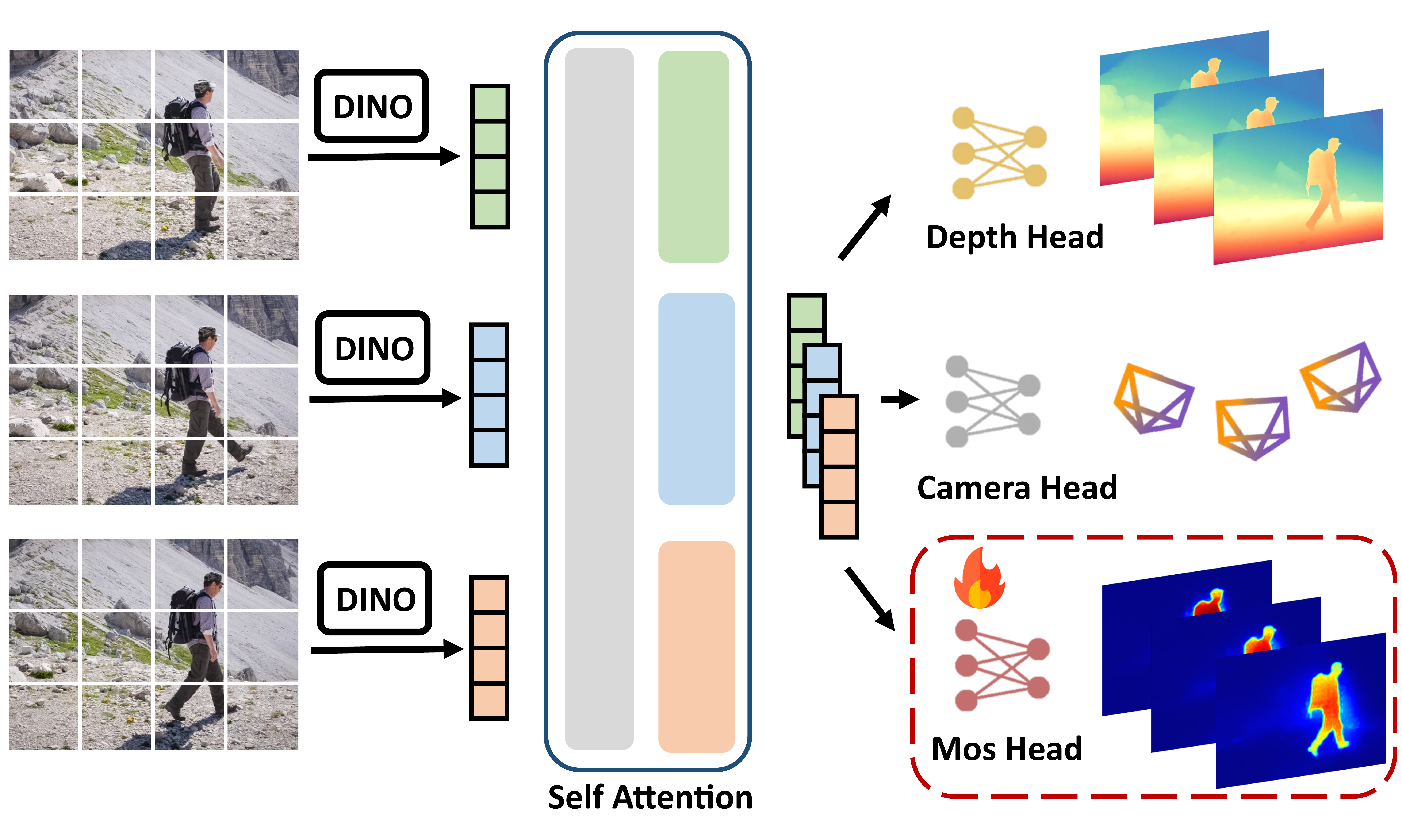}
    \caption{Network architecture of $\pi^3_{\text{mos}}$. After patchifying input images into tokens using DINOv2, we apply alternating frame-wise and global self-attention. The fused tokens are fed into three prediction heads: a depth head for estimating depth, a camera head for predicting relative poses, and a Moving object segmentation~(MOS) head for producing motion probabilities. We highlight the MOS head as the key addition compared to $\pi^3$.} \label{Fig:pi3}
    \vspace{-0.3cm}
\end{figure}

We further optimize $\mathcal{L}_{\text{BA}}$ using the Levenberg-Marquardt (LM) algorithm to jointly estimate the camera pose of each frame and the depth of each patch within the sliding window.
After linearizing \cref{eq:BA}, the corresponding normal equations can be formulated as follows:
\begin{equation}
\begin{bmatrix} 
\mathbf{B} & \mathbf{E} \\ 
\mathbf{E}^\top & \mathbf{C} 
\end{bmatrix} 
\begin{bmatrix} \Delta\boldsymbol{\xi} \\\Delta \mathbf{d} \end{bmatrix} = \begin{bmatrix} \mathbf{v}\\\mathbf{w} \end{bmatrix},
\label{eq:normal}
\end{equation}
where $\mathbf{B}$ and $\mathbf{C}$ are the Hessian matrices for frame poses and patch inverse depths, $\mathbf{E}$ is the cross-coupling matrix, $\Delta\boldsymbol{\xi}$ and $\Delta\mathbf{d}$ are the pose and depth increments, and $\mathbf{v}$, $\mathbf{w}$ are the corresponding gradients.
This linear equation can be efficiently solved using the Schur complement:
\begin{align}
  \Delta\boldsymbol{\xi} &= \left[\mathbf{B} - \mathbf{E}\mathbf{C}^{-1}\mathbf{E}^\top\right]^{-1}\left(\mathbf{v} - \mathbf{E}\mathbf{C}^{-1}\mathbf{w}\right) \label{eq:schur_pose} \\
  \Delta\mathbf{d} &= \mathbf{C}^{-1}\left(\mathbf{w} - \mathbf{E}^\top\Delta\boldsymbol{\xi}\right) \label{eq:schur_depth}
\end{align}

\begin{figure*}[t]
  \centering
  \includegraphics[width=1\linewidth]{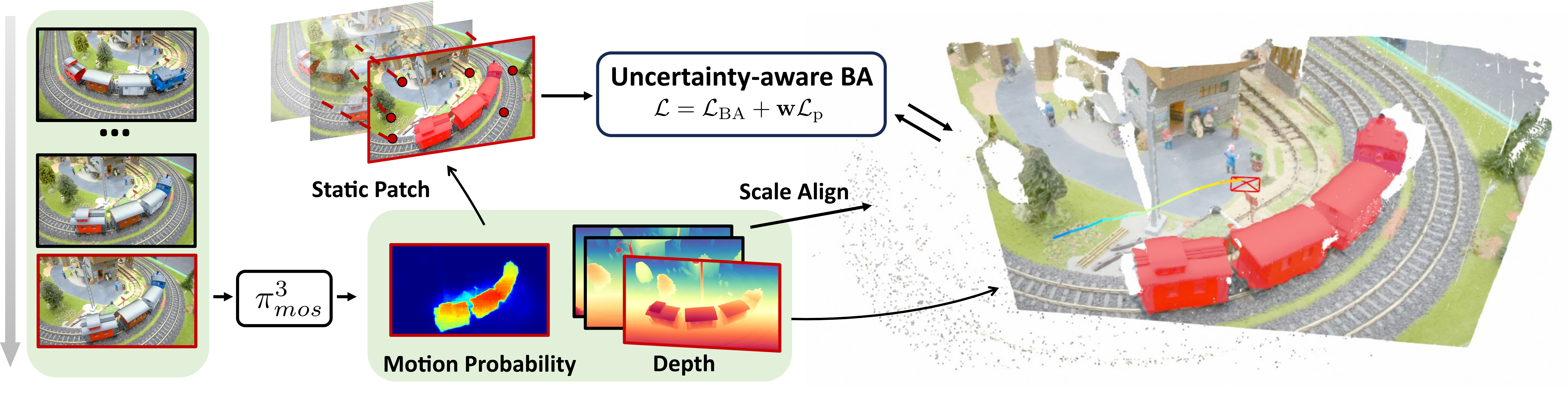}
  \caption{Overview of our SLAM system. We highlight the current frame with a red outline. It is processed together with a set of selected historical frames by our $\pi^3_{\text{mos}}$ model, which outputs per-frame motion probabilities and depth predictions. We use the current frame’s motion probabilities to initialize new patches only on the static background.
  The depth maps of the historical frames are used to align the scale of the current prediction via patch registration. After scale alignment, the refined depth and the optical flow constraints are fed into our BA module to jointly optimize camera poses and patch depths.}
  \label{fig:overview}
  \vspace{-0.3cm}
\end{figure*}

\subsection{Moving Object Segmentation by $\pi^3_{\text{mos}}$}
\label{subsec:mos}
Our $\pi^3_{\text{mos}}$ extends original $\pi^3$~\cite{wang2025arxiv-pspe} while preserving its core architecture. As shown in \cref{Fig:pi3}, %by adding a new head for moving object segmentation
given a set of images, $\pi^3_{\text{mos}}$ first extracts semantic features from each image using \mbox{DINOv2~\cite{oquab2024tmlr}} and then employs multiple layers of alternating frame-wise and global self-attention to fuse information across images. 
Finally, three lightweight decoders are used to regress the relative camera pose, depth map and pixel-wise motion probability for each image.

% Due to the inclusion of extensive dynamic-scene data during training, $\pi^3$ demonstrates stable performance even in highly dynamic environments.
% However, since $\pi^3$ does not explicitly segment moving objects, aligning the output of these feed-forward reconstruction models to handle long sequences remains challenging.

% Inspired by the strong performance of $\pi^3$ in 3D reconstruction, we hypothesize that moving object masks can be effectively derived from its feature representations, which encode rich semantic and geometric information.
% Therefore, we extend the original $\pi^3$ architecture by adding a moving objects segmentation head that predicts the motion probability for each pixel.
% We denote the extended model as $\pi^3_{\text{mos}}$ to distinguish it from the original version.
More specifically, given an input batch of RGB images   
$\mathcal{I} = \{\mathbf{I}_i \in \mathbb{R}^{H\times W\times 3}\}_{i=1}^{N}$,
the predictions of $\pi^3_{\text{mos}}$ are:
\begin{align}
  \pi^3_{\text{mos}}(\mathcal{I})&=\left ( \mathbf{T}_i, \mathbf{D}_i, \mathbf{C}_i, \mathbf{M}_i\right ) _{i=1}^N,
 \label{eq:pi3}
\end{align}
where $\mathbf{T}_i \in SE(3)$ denotes the camera pose, $\mathbf{D}_i \in \mathbb{R}^{H\times W}$ is the dense depth image derived from the predicted point cloud with corresponding confidence scores $\mathbf{C}_i\in \mathbb{R}^{H\times W}$,
and $\mathbf{M}_i \in [0,1]^{H\times W}$ represents the per-pixel motion probability.
Note that $\mathbf{D}_i$ is defined only up to an unknown scale, necessitating an additional scale estimation step, described in \cref{subsec:scale}, to ensure consistent scaling.

\subsection{System Overview}
\label{subsec:system}
As illustrated in \cref{fig:overview}, once the current frame $i$ arrives, it is paired with a set of selected historical keyframes and jointly processed by $\pi^3_{\text{mos}}$ to predict the per-pixel motion probability $\mathbf{M}_i$ and depth map $\mathbf{D}_i$.
We then apply a threshold $s_d$ to identify pixels with $\mathbf{M}_i < s_d$ as the static region, from which we randomly sample $K$ patches to compute optical flow with respect to historical frames, thereby establishing inter-frame geometric constraints. 
This ensures that subsequent pose optimization is unaffected by moving objects.
We further align the predicted depth $\mathbf{D}_i$ with the sparse point cloud formed by previously optimized patches to resolve the scale ambiguity of $\pi^3_{\text{mos}}$ predictions and ensure consistency with the system scale.
Finally, the aligned depth and the established flow constraints are added into the sliding window, where an uncertainty-aware bundle adjustment jointly optimizes the camera poses and patch depths.

\subsection{Depth Scale Estimation}
\label{subsec:scale}
We observe that the predicted depth scales remain consistent across frames within the same input batch, but this consistency does not hold across different batches.
For instance, when the same image is paired with different images to form separate input batches for $\pi^3_{\text{mos}}$ inference, its predicted depth scale may vary.
This necessitates aligning the depth predictions from $\pi^3_{\text{mos}}$ with a constant scale to enable incremental 3D reconstruction.

% scale-consistent depths estimated by our SLAM system before incorporating them as priors in the bundle adjustment.
To address this scale ambiguity, we select $N-1$ historical keyframes and feed them, together with the current frame, into $\pi^3_{\text{mos}}$ to obtain the corresponding depth maps $\{\mathbf{D}_i\}_{i=1}^{N}$ and confidence maps $\{\mathbf{C}_i\}_{i=1}^{N}$ as described in \cref{eq:pi3}, where $\mathbf{D}_N$ corresponds to the depth map of the current frame.
Since we do have a reference scale for the current frame, we refer to historical keyframes, whose scales have been determined and consistent with each other.
For each historical frame processed by $\pi^3_{\text{mos}}$, we sample $\mathbf{D}_i$ and $\mathbf{C}_i$ at the center location of $P_k^i$ to obtain the corresponding inverse depths $\hat{d}_k^i = 1/\hat{z}_k^i$ and confidence scores $c_k^i$.
We want to estimate a single scale factor $s$ in the inverse-depth domain that aligns the newly predicted depth from all selected historical frames $\{\mathbf{D}_i\}_{i=1}^{N-1}$ with their previously estimated inverse depths of patches.
To this end, we optimize the scale $s$ using:
\begin{equation}
  s^* = \arg\min_s \sum_{i=1}^{N-1} \sum_{k=1}^K c_k^i \cdot \rho\left(d_k^i - s \cdot \hat{d}_k^i\right),
\label{eq:huber_irls}
\end{equation}
where $\rho(\cdot)$ denotes the Huber loss function, and $d_k^i$ represents the already scaled depth of patch $P_k^i$ in a historical frame.
We initialize $s$ using the weighted median of the ratio $d_k / d_k^p$, where the weights are derived from the confidence scores provided by $\pi^3_{\text{mos}}$. 
This initialization yields a robust starting point that is less sensitive to outliers than a simple mean.
Finally, we scale the current depth map $\mathbf{D}_N$ with the optimized scale value $s^*$, thereby achieving consistent scaling during online mapping.
Since the patches are selected exclusively from static regions, the scale estimation remains unaffected by moving objects.

\subsection{Uncertainty-aware BA}
\label{subsec:weighting}

In this section, we describe how the bundle adjustment module integrates optical flow constraints with the predicted depths from the feed-forward reconstruction model.
We observe that when optical flow tracking is reliable and the camera exhibits sufficient translational motion, solving bundle adjustment can effectively estimate camera poses and inverse depth of patches.
However, when the camera's translational motion is insufficient, relying solely on optical flow constraints to compute patch depths introduces significant uncertainty, making it difficult to accurately recover the geometric structure of the environment. 
Introducing priors from a 3D feed-forward reconstruction model at this stage can improve the robustness of state estimation.

To incorporate depth predictions from $\pi^3_{\text{mos}}$ as priors, we introduce a depth prior loss:
\begin{equation}
  \mathcal{L}_{\text{p}} = \sum_{f=1}^{F} \sum_{k=1}^{K}\left \| d_k^f - s_f\hat{d}_k^f \right \|^2,
 \label{eq:prior}
\end{equation}
where $F$ is the number of frames in the optimization window, $K$ is the number of patches per frame,  $\hat{d}_k^f$ is the inverse depth of patch $P_k^f$ predicted by $\pi^3_{\text{mos}}$, and $s_f$ is the scale factor for frame $f$ that ensures scale consistency of the depth predictions.
The final loss function~$\mathcal{L}$ to be optimized combines the bundle adjustment loss~$\mathcal{L}_{\text{BA}}$ with the depth prior loss~$\mathcal{L}_{\text{p}}$ as follows:
\begin{align}
  \mathcal{L} &= \mathcal{L}_{\text{BA}} + \mathbf{w} \mathcal{L}_{\text{p}} \\
              &= \mathcal{L}_{\text{BA}} + \sum_{f=1}^{F} w_f \sum_{k=1}^{K}\left \| d_k - s_f\hat{d}_k^f \right \|^2,
\end{align}
where $\mathbf{w} = [w_1, w_2, \ldots, w_F]^\top$ is a weight vector. To enhance camera tracking stability using depth priors while preventing accuracy degradation caused by over-reliance on them, we introduce an adaptive weighting strategy based on the uncertainty of state estimation, inspired by MegaSaM~\cite{li2025cvpr-mafa}.
Specifically, when the uncertainty is high, we increase the weight $w_f$ to stabilize the optimization. When the uncertainty is low, we reduce its influence and rely more on the original BA optimization. 
We quantify uncertainty by computing the variance of patch depth estimates.
Following NeRF-SLAM~\cite{rosinol2023wacv,rosinol2022}, the marginal covariance of the inverse depths $\mathbf{\Sigma_d}$ can be derived from \cref{eq:normal} as
\begin{equation} 
  \mathbf{\Sigma_d} = \mathbf{C}^{-1} + \mathbf{C}^{-1} \mathbf{E}^\top \Sigma_T \mathbf{E} \mathbf{C}^{-1}, 
  \label{eq:cov_depth_new}
\end{equation}  
where
\begin{equation}
  \mathbf{\Sigma_T} = \bigl(\mathbf{B} - \mathbf{E} \mathbf{C}^{-1} \mathbf{E}^\top \bigr)^{-1}
\end{equation}  
denotes the covariance of the camera poses, i.e., the inverse of the Schur complement of the full Hessian in \cref{eq:normal} with respect to $\mathbf{C}$. This matrix inversion can be efficiently computed via Cholesky decomposition.
The variance $\sigma_{d_j}^2$ of an individual inverse depth $d_j$ is
\begin{equation}
  \sigma_{d_j}^2 = \bigl[\mathrm{diag}(\mathbf{\Sigma_d})\bigr]_j,
\end{equation}  
where $j$ indexes the corresponding patch in the optimization window. Using standard nonlinear uncertainty propagation, the standard deviation of the actual depth $z_j = 1/d_j$ is $\sigma_{z_j} = \sigma_{d_j}/{d_j^2}.$
To achieve scale-invariant uncertainty, we define the relative standard deviation as
\begin{equation}
  \sigma_{z_j}^{\mathrm{rel}} = \frac{\sigma_{z_j}}{z_j} = \frac{\sigma_{d_j}}{d_j}.
\end{equation}

Finally, the frame-wise uncertainty is summarized by the median relative standard deviation $\sigma_{z,\mathrm{med}}^{\mathrm{rel}}$, which is mapped to a frame weight via a sigmoid function:
\begin{equation}
  w_f = 1/(1 + \exp(-\alpha(\sigma_{z,\mathrm{med}}^{\mathrm{rel}} - \beta))),
\end{equation}
where $\alpha$ and $\beta$ are hyperparameters that control the steepness and the offset of the sigmoid curve.

In addition, during scale estimation in \cref{subsec:scale}, we remove points whose relative standard deviation $\sigma_{z_j}^{\mathrm{rel}}$ exceeds a threshold $t_{\sigma}$ to improve the stability of scale estimation.

%%%%%%%%%%%%%%%%%%%%%%%%%%%%%%%%%%%%%%%%%%%%%%%%%%%%%%%%%%%%%%%%
\subsection{Implementation Details}
\label{subsec:Implementation}
We trained our $\pi^3_{\text{mos}}$ model using four datasets, including three synthetic ones (Kubric~\cite{greff2021cvpr-kasd}, Dynamic Replica~\cite{karaev2023cvpr}, and Virtual KITTI 2~\cite{cabon2020arxiv}) and one real-world indoor dataset (HOI4D~\cite{liu2022cvpr_ha4e}).
These datasets either provide ground-truth moving object masks directly or contain scene flow annotations from which such masks can be derived.
During training, we use the weights from the original $\pi^3$ as the start-point and only train the newly introduced MOS head. The motion probability prediction is supervised using a simple binary cross-entropy~(BCE) loss.
% Since all other parameters were frozen during training, we only use the raw RGB images and corresponding moving object masks, without requiring depth or camera pose information.
In total, we sampled approximately 250 k images from HOI4D, 120 k from Kubric, 14.5 k from Dynamic Replica, and 21 k from Virtual KITTI~2.
Training was performed on 8 NVIDIA A40 GPUs for 48 hours in total.
We provide the hyperparameter settings of our SLAM system in the supplementary.
% For the SLAM system, we set the number of patches per frame to $K=128$, and the threshold applied to the motion probability for distinguishing static and dynamic regions to $s_d=0.4$.
% The sliding window size for pose optimization is set to 10. The threshold $t_{\sigma}$, which is used to reject unreliable patches in scale estimation, is 0.75. In each time, 5 historical keyframes together with the current frame are fed into $\pi^3_{\text{mos}}$ for inference.
\begin{table*}[t]
  \centering
  \caption{Camera tracking performance on the Bonn \mbox{RGB-D} Dynamic Dataset~\cite{palazzolo2019iros} (ATE RMSE $\downarrow$ [cm]).}
  \label{tab:tracking_performance_bonn}
  \small
  \resizebox{0.85\textwidth}{!}{%
  \begin{tabular}{l|cccccccc|c}
  \hline
  \textbf{Method} & \texttt{Balloon} & \texttt{Balloon2} & \texttt{Crowd} & \texttt{Crowd2} & \texttt{Person} & \texttt{Person2} & \texttt{Moving} & \texttt{Moving2} & \textbf{Avg.} \\
  \hline
  \rowcolor{gray!13}
  \multicolumn{10}{l}{\textit{\mbox{RGB-D} Methods}} \\
  Refusion~\cite{palazzolo2019iros} & 17.5 & 25.4 & 20.4 & 15.5 & 28.9 & 46.3 & 7.1 & 17.9 & 22.38 \\
  ORB-SLAM2~\cite{mur-artal2017tro} & 6.5 & 23.0 & 4.9 & 9.8 & 6.9 & 7.9 & 3.2 & 3.9 & 6.36 \\
  DynaSLAM~\cite{bescos2018ral} & 3.0 & 2.9 & 1.6 & 3.1 & 6.1 & 7.8 & 23.2 & 3.9 & 6.45 \\
  % DG-SLAM~\cite{wang2022icra} & 3.7 & 4.1 & - & - & 4.5 & 6.9 & - & 3.5 & - \\
  % RoDyn-SLAM~\cite{wang2023icra} & 7.9 & 11.5 & - & - & 14.5 & 13.8 & - & 12.3 & - \\
  DDN-SLAM~\cite{li2025ral-dsrt} & 1.8 & 4.1 & 1.8 & 2.3 & 4.3 & 3.8 & 2.0 & 3.2 & 2.91 \\
  \hline
  \rowcolor{gray!13}
  \multicolumn{10}{l}{\textit{Monocular Methods}} \\
  % DSO~\cite{engel2018pami} & 7.3 & 21.8 & 10.1 & 7.6 & 30.6 & 26.5 & 4.7 & 11.2 & - & - & 15.0 \\
  DROID-SLAM~\cite{teed2021neurips} & 7.5 & 4.1 & 5.2 & 6.5 & 4.3 & 5.4 & 2.3 & 4.0 & 4.91 \\
  % DynaMoN (MS)~\cite{wang2024cvpr} & 6.8 & 3.8 & 6.1 & 5.6 & 2.4 & 3.5 & 1.4 & 2.6 & 4.02 \\
  % DynaMoN (MS\&SS)~\cite{wang2024cvpr} & 2.8 & 2.7 & 3.5 & 2.8 & 14.8 & 2.2 & 1.3 & 2.7 & 4.10 \\
  MonST3R~\cite{zhang2025iclr} & 5.4 & 7.2 & 5.4 & 6.9 & 11.9 & 11.1 & 3.3 & 7.4 & 7.3 \\
  MegaSaM~\cite{li2025cvpr-mafa} & 3.7 & 2.6 & 1.6 & 7.2 & 4.1 & 4.0 & 1.4 & 3.4 &3.51 \\
  WildGS-SLAM~\cite{zheng2025cvpr-wsmg} & \underline{2.8} & \underline{2.4} & \underline{1.5} & \underline{2.3} & \textbf{3.1} & \textbf{2.7} & \underline{1.6} & \underline{2.2} & \underline{2.36} \\
  \midrule
  \textbf{ours} & \textbf{2.6} & \textbf{2.2} & \textbf{1.3} & \textbf{2.2} & \underline{3.2} & \underline{3.0} & \textbf{1.2} & \textbf{1.9} & \textbf{2.20} \\
  \hline
  \end{tabular}%
  }
  \vspace{-0.3cm}
\end{table*}

\begin{table}[t]
  \centering
  \caption{Camera tracking performance on the Wild-SLAM MoCap Dataset~\cite{zheng2025cvpr-wsmg} (ATE RMSE $\downarrow$ [cm]).}
  \label{tab:tracking_performance_wild}
  \normalsize
  \resizebox{0.475\textwidth}{!}{%
  \begin{tabular}{l|ccccc|c}
  \hline
  \textbf{Method} & \texttt{Person} & \texttt{Racket} & \texttt{Table1} & \texttt{Table2} & \texttt{Umbre.} & \textbf{Avg.} \\
  \hline
  \rowcolor{gray!15}
  \multicolumn{7}{l}{\textit{\mbox{RGB-D} Methods}} \\
  Refusion & 5.0 & 10.4 & 99.1 & 101.0 & 10.7 & 45.24 \\
  DynaSLAM & 0.5 & 0.8 & 1.2 & 34.8 & 34.7 & 14.40 \\
  \hline
  \rowcolor{gray!15}
  \multicolumn{7}{l}{\textit{Monocular Methods}} \\
  % DSO~\cite{engel2018pami} & 12.0 & 2.5 & 1.0 & 88.6 & 9.3 & 3.1 & 41.5 & 50.6 & 85.3 & 26.0 & 32.99 \\
  DROID-SLAM & 0.6 & 1.5 & 48.0 & 95.6 & 3.8 & 29.90 \\
  DynaSLAM & \underline{0.4} & \underline{0.6} & 1.8 & 42.1 & 1.2 & 9.22 \\
  MonST3R & 7.2 & 13.2 & 4.8 & 33.7 & 5.5 & 12.88 \\
  MegaSaM & 3.2 & 1.6 & 1.0 & 9.4 & 0.6 & 3.16 \\
  WildGS-SLAM & 0.8 & \textbf{0.4} & \underline{0.6} & \underline{1.3} & \textbf{0.2} & \underline{0.66} \\
  \midrule
  \textbf{ours} & \textbf{0.2} & \textbf{0.4} & \textbf{0.2} & \textbf{1.1} & \textbf{0.2} & \textbf{0.42} \\
  \hline
  \end{tabular}%
  }
  \vspace{-0.4cm}
\end{table}

%%%%%%%%%%%%%%%%%%%%%%%%%%%%%%%%%%%%%%%%%%%%%%%%%%%%%%%%%%%%%%%%
\section{Experiments}
\label{sec:experiments}
% In this section, we demonstrate the effectiveness of our proposed approach through comprehensive evaluation including:
% (1) \textbf{Moving object segmentation} to support our first contribution that our proposed $\pi^3_{\text{mos}}$ model can accurately segment moving objects.
% (2) \textbf{Camera tracking in dynamic datasets} to demonstrate our approach have superior camera pose estimation accuracy on dynamic scene datasets, achieving the highest average tracking accuracy compared to both \mbox{RGB-D} and monocular SLAM baselines.
% (3) \textbf{Video depth estimation}: Our method produces accurate and scale-consistent depth predictions, showing significant improvements over feed-forward reconstruction models on standard depth estimation benchmarks.
In this section, we evaluate our method against state-of-the-art approaches on multiple tasks to demonstrate its effectiveness and validate our contributions.
For all the tables in this section, we highlight the best results in \textbf{bold} and the second best results are \underline{underscored}.

\subsection{Moving Object Segmentation}
\label{subsec:dynamicOS}
\textbf{Datasets.} We first discuss the moving object segmentation module in our pipeline, which serves as a crucial component of the SLAM system.
Following the experimental setup in Easi3R, we conduct experiments on the moving object segmentation benchmarks DAVIS-16~\cite{perazzi2016cvpr} and the more challenging DAVIS-17~\cite{pont-tuset2017arxiv}.
Since our goal is online robotic applications, we directly evaluate the performance of moving object segmentation from each model's raw output, without applying any time-consuming post-processing.

\textbf{Metric and Baselines.} We report performance using the IoU mean (JM) and IoU recall (JR) metrics.
JM measures the overall segmentation quality, while JR reflects segmentation robustness, representing the percentage of frames with IoU greater than 50 across the entire sequence.
We compare our $\pi^3_{\text{mos}}$ with several state-of-the-art pose-free 4D reconstruction methods, including DUSt3R~\cite{wang2024cvpr-dg3v}, MonST3R~\cite{zhang2025iclr}, DAS3R~\cite{xu2024arxiv-ddag}, and Easi3R~\cite{chen2025iccv} using DUSt3R and MonST3R as the backbone.
All these methods are built upon the DUSt3R framework and extend it to handle dynamic video inputs.

\textbf{Results.}
We present the qualitative results in \cref{tab:davis}.
DUSt3R and MonST3R rely on optical flow estimation to distinguish moving objects, which leads to degraded performance on textureless scenes.
DAS3R extends DUSt3R model by fine-tuning on dynamic datasets, but yields only minor gains on DAVIS-17 and even worse performance on DAVIS-16 benchmark.
Easi3R demonstrates promising results by utilizing attention maps from DUSt3R or MonST3R model without any fine-tuning on dynamic data.
However, all these DUSt3R-based models are limited by the lack of comprehensive spatial context, as DUSt3R is originally designed for image-pair inputs.
In contrast, our feed-forward reconstruction model $\pi^3_{\text{mos}}$ is trained for multi-view input and also enhanced with general semantic features provided by \mbox{DINOv2~\cite{oquab2024tmlr}}.
% We exploit the strong prior knowledge embedded in the $\pi^3$ model to train a dedicated moving object segmentation head, which produces high-quality moving object masks.
Across both benchmarks, our approach achieves the best performance in terms of both segmentation quality and robustness, with a substantial margin over all baseline approaches.

\begin{figure}[!t]
  \centering
  \begin{subfigure}[t]{0.242\columnwidth}
      \includegraphics[width=\linewidth]{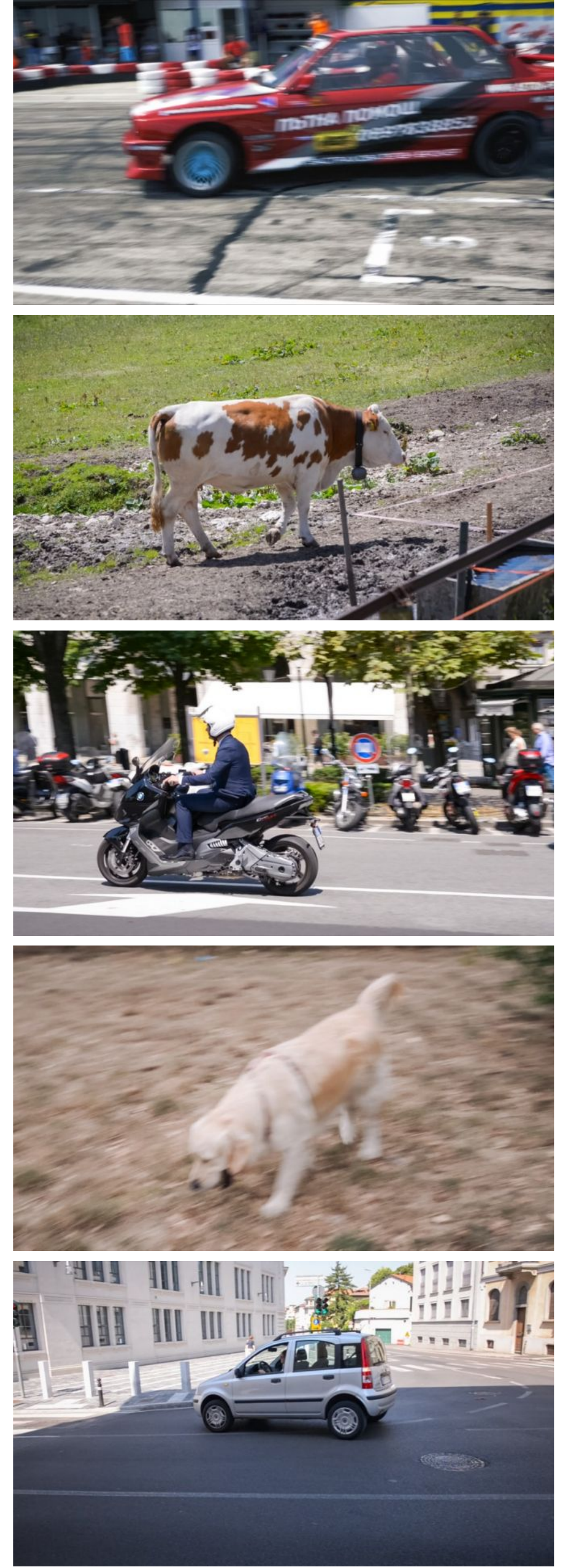}
      \caption{Input}
  \end{subfigure}
  % \begin{subfigure}[t]{0.19\columnwidth}
  %     \includegraphics[width=\linewidth]{pics/davis-column2.pdf}
  %     \caption{\footnotesize Easi3R$_{\text{DUSt3R}}$~\cite{chen2025iccv}}
  % \end{subfigure}
  \begin{subfigure}[t]{0.242\columnwidth}
      \includegraphics[width=\linewidth]{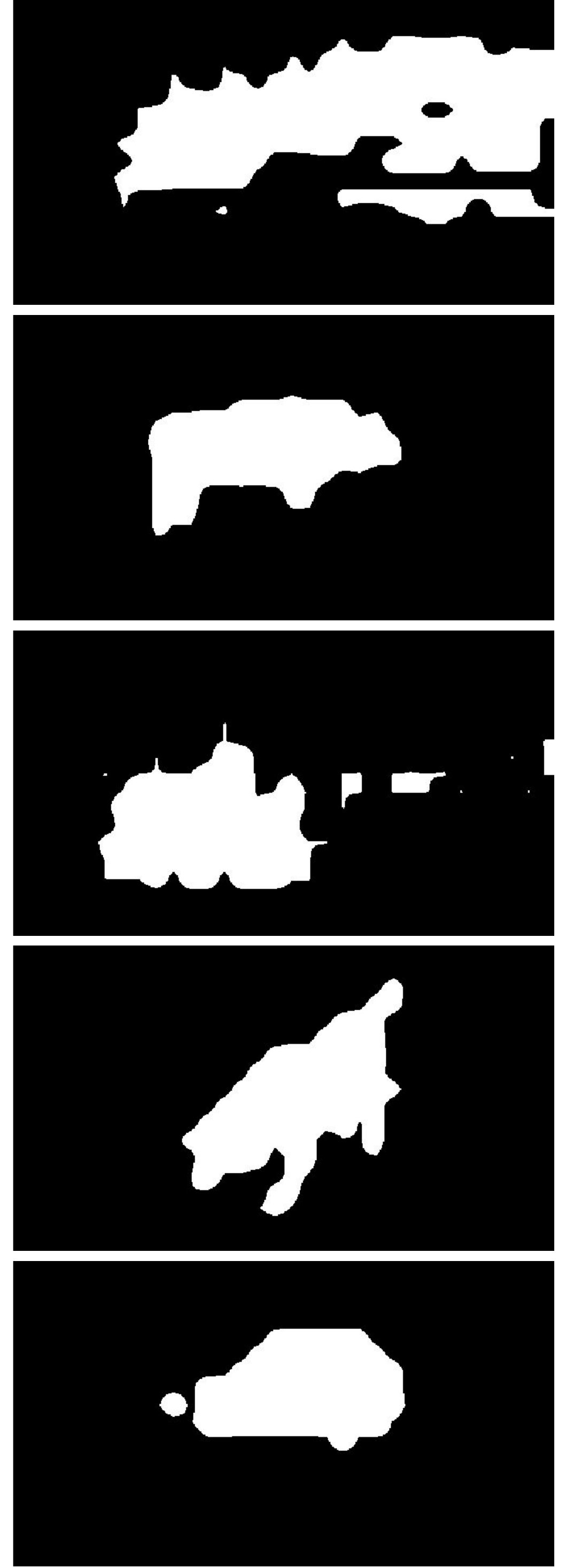}
      \caption{Easi3R~\cite{chen2025iccv}}
  \end{subfigure}
  \begin{subfigure}[t]{0.242\columnwidth}
      \includegraphics[width=\linewidth]{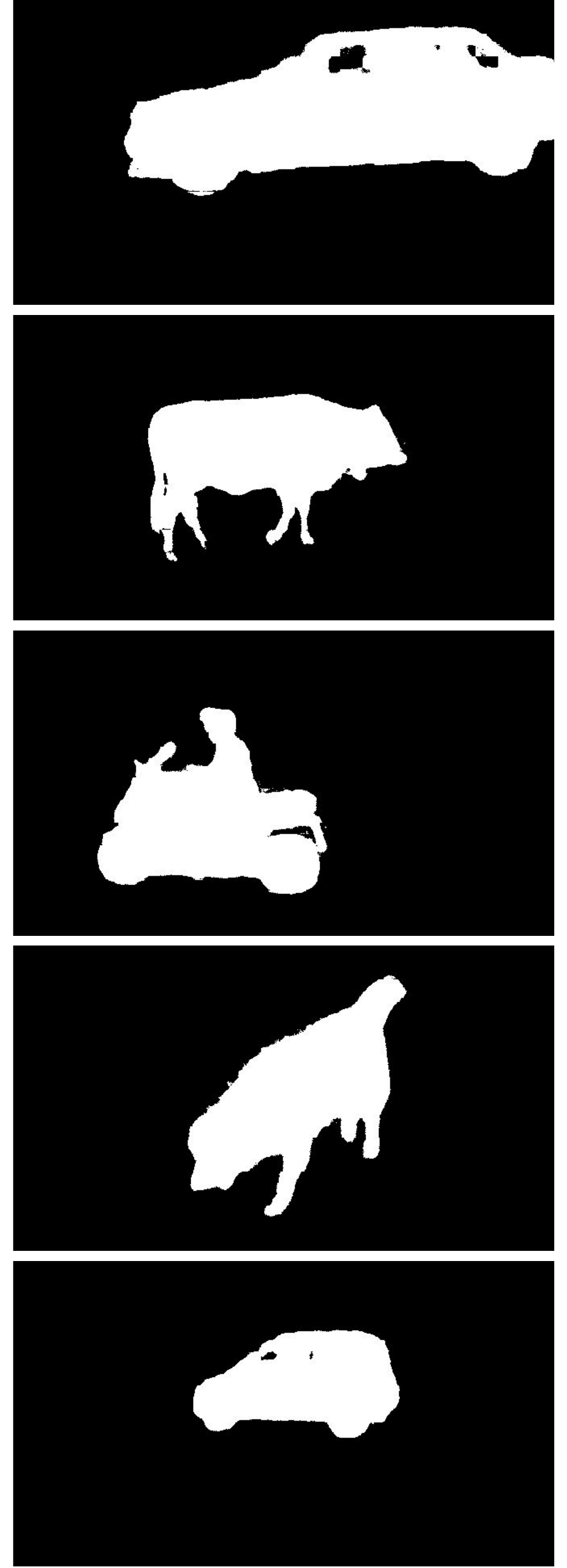}
      \caption{Ours}
  \end{subfigure}
  \begin{subfigure}[t]{0.242\columnwidth}
      \includegraphics[width=\linewidth]{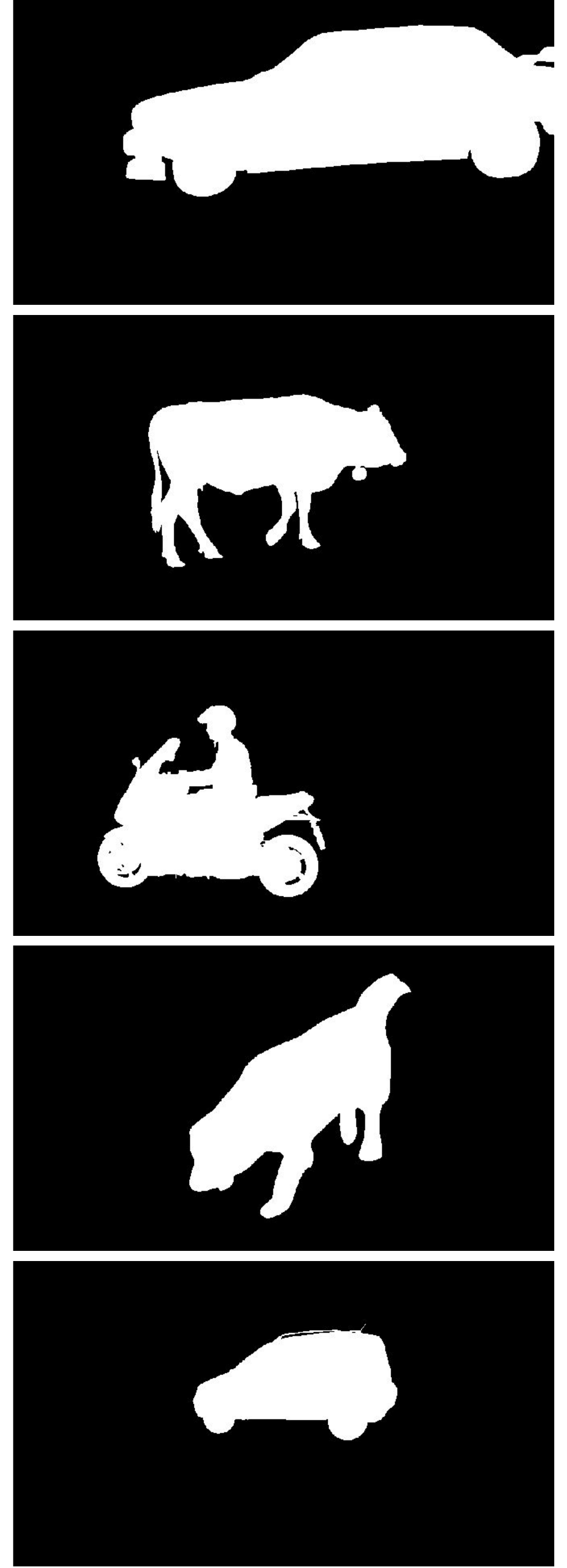}
      \caption{GT}
  \end{subfigure}
  \caption{Qualitative comparison of moving object segmentation on the DAVIS-16 dataset.
  We show the moving object masks predicted from Easi3R$_{\text{MonST3R}}$ and our $\pi^3_{\text{mos}}$ model.
  Our method produces more accurate segmentation results, particularly in challenging scenarios with motion blur and thin structures. 
  }
  \label{F:davis-visual}
  \vspace{-0.4cm}
\end{figure}

We further illustrate the quantitative results in \cref{F:davis-visual}.
Our approach produces precise moving object masks, even in challenging scenarios with motion blur or thin structures in the input frames.
These results highlight the model's capabilities to preserve fine spatial details and maintain segmentation stability under real-world dynamic conditions.

% single column
% \begin{figure}[!t]
%     \centering
%     \includegraphics[width=0.99\columnwidth]{pics/davis-visual.pdf}
%     \caption{xxxx} \label{F:davis-visual}
% \end{figure}
% single column

\begin{table}[t]
  \centering
  \caption{Moving object segmentation performance on DAVIS-16 and DAVIS-17 benchmarks.}
  \label{tab:davis}
  \small
%   \setlength{\tabcolsep}{4pt}
%   \renewcommand{\arraystretch}{1.15}
%   \resizebox{0.9\columnwidth}{!}
{%
  \begin{tabular}{l|cc|cc}
  \hline
  \multirow{2}{*}{\textbf{Method}} & 
  \multicolumn{2}{c|}{\texttt{DAVIS-16}} & 
  \multicolumn{2}{c}{\texttt{DAVIS-17}} \\
  \cmidrule(lr){2-3}\cmidrule(lr){4-5}
   & JM$\uparrow$ & JR$\uparrow$ & JM$\uparrow$ & JR$\uparrow$ \\
  \hline
  DUSt3R~\cite{wang2024cvpr-dg3v}      & 42.1 & 45.7 & 35.2 & 35.3 \\
  MonST3R~\cite{zhang2025iclr}         & 40.9 & 42.2 & 38.6 & 38.2 \\
  DAS3R~\cite{xu2024arxiv-ddag}        & 41.6 & 39.0 & 43.5 & 42.1 \\
  Easi3R$_{\text{DUSt3R}}$~\cite{chen2025iccv}  & 53.1 & 60.4 & 49.0 & 56.4 \\
  Easi3R$_{\text{MonST3R}}$~\cite{chen2025iccv} & \underline{57.7} & \underline{71.6} & \underline{56.5} & \underline{68.6}  \\
  \midrule
  \textbf{ours}        & \textbf{68.1} & \textbf{78.3} & \textbf{70.6} & \textbf{81.3} \\
  \hline
  \end{tabular}}
  \vspace{-0.5cm}
\end{table}

\subsection{Camera Tracking}
\label{subsec:tracking}
\textbf{Datasets.} We evaluate the camera tracking performance of our SLAM system on three dynamic scene datasets.
Among them, two are real-world datasets, the Bonn \mbox{RGB-D} Dynamic Dataset~\cite{palazzolo2019iros} and the Wild-SLAM MoCap Dataset~\cite{palazzolo2019iros}.
Both of these two datasets are recorded in single-room indoor environments and provide accurate ground-truth camera poses obtained using a motion capture system.
They contain various types of moving objects beyond humans, such as tables, laptops, and umbrellas, which enable a comprehensive evaluation of the robustness of dynamic SLAM algorithms in different scenarios.
The Bonn \mbox{RGB-D} sequences range from 400--1000 frames, and those in the Wild-SLAM dataset mostly contain 1000--2000 frames. 
In addition, we further use the MPI Sintel dataset~\cite{butler2012eccv} to evaluate our system under more challenging conditions.
Sintel contains sequences rendered from 3D animated movies, covering large-scale outdoor scenes with significant depth variation, complex object motion, and fast camera movement with motion blur.
We select 14 sequences from this dataset, each containing 20--50 frames.

\textbf{Metric and Baselines.} For the Bonn~\cite{palazzolo2019iros} and Wild-SLAM~\cite{zheng2025cvpr-wsmg} datasets, we report the Absolute Trajectory Error~(ATE) for each sequence, which is computed as the root mean square error~(RMSE) of all poses after alignment.
For the Sintel dataset, following the evaluation setup in~\cite{zhang2022eccv-samf, chen2024cvpr-lvlt,chen2025iccv-botb},  we report the average ATE, Relative Translation Error~(RTE), and Relative Rotation Error~(RRE) of all selected sequences.
The RTE and RRE are computed as the mean translation and rotation differences between consecutive estimated and ground-truth poses over all frame pairs.
The baselines include SLAM systems that do not explicitly handle moving objects, ORB-SLAM2~\cite{mur-artal2017tro}, DROID-SLAM~\cite{teed2021neurips}, and TartanVO~\cite{tartanvo2020corl};
methods that rely on closed-world deep learning-based object detection, DynaSLAM~\cite{bescos2018ral} (which has both RGB-D and RGB versions) and DDN-SLAM~\cite{li2025ral-dsrt}; a geometry-based approach Refusion~\cite{palazzolo2019iros}; feed-forward reconstruction model MonST3R~\cite{zhang2025iclr};
methods that distinguish static and dynamic tracks via long-term point tracking, including LEAP-VO~\cite{chen2024cvpr-lvlt} and BA-Track~\cite{chen2025iccv-botb}; WildGS-SLAM~\cite{zheng2025cvpr-wsmg} and offline MegaSaM~\cite{li2025cvpr-mafa} approach. For MegaSaM, we report its performance from segment-aligned version implemented in WildGS-SLAM~\cite{zheng2025cvpr-wsmg}.

\begin{table}[t]
  \centering
  \caption{Camera tracking performance on the Sintel~\cite{butler2012eccv} Dataset.}
  \label{tab:tracking_performance_sintel}
  \small
  \renewcommand{\arraystretch}{1.0}
  \resizebox{\columnwidth}{!}{%
  \begin{tabular}{l|ccc}
  \hline
  \textbf{Method} & ATE (cm)$\downarrow$ & RTE (cm)$\downarrow$ & RRE (deg)$\downarrow$ \\
  \midrule
  % Refusion~\cite{palazzolo2019iros} & 4.2 & 5.6 & 5.0 & 91.9 & 5.0 & 10.4 & 39.4 & 99.1 & 101.0 & 10.7 & 37.23 \\
  % DynaSLAM (RGBD)~\cite{bescos2018ral} & 1.6 & 0.5 & \underline{0.5} & 1.7 & 0.5 & 0.8 & 2.1 & 1.2 & 34.8 & 34.7 & 7.84 \\
  % \hline
  % \rowcolor{gray!15}
  % \multicolumn{12}{l}{\textit{Monocular Methods}} \\
  % DSO~\cite{engel2018pami} & 12.0 & 2.5 & 1.0 & 88.6 & 9.3 & 3.1 & 41.5 & 50.6 & 85.3 & 26.0 & 32.99 \\
  DROID-SLAM~\cite{teed2021neurips} & 17.5 & 8.4 & 1.91 \\
  TartanVO~\cite{tartanvo2020corl} & 23.8 & 9.3 & 1.31 \\
  DytanVO~\cite{shen2023icra-djro} &13.1 & 9.7 &1.54 \\
  MonST3R~\cite{zhang2025iclr} & 7.8 & 3.8 & 0.49 \\
  DPVO~\cite{teed2023nips} & 11.5 & 7.2 & 1.98 \\
  LEAP-VO~\cite{chen2024cvpr-lvlt} &8.9 & 6.6 & 1.25 \\
  BA-Track~\cite{chen2025iccv-botb} &\underline{3.4} &\underline{2.3} & \underline{0.12} \\
  WildGS-SLAM~\cite{zheng2025cvpr-wsmg} &18.2 & 9.4 & 1.57\\
  \midrule
  \textbf{ours}  & \textbf{1.9} & \textbf{1.0} & \textbf{0.11} \\
  \hline
  \end{tabular}%
  }
  \vspace{-0.3cm}
\end{table}

\begin{figure}[!t]
  \centering
  \begin{subfigure}[b]{\columnwidth}
      \centering
      \includegraphics[width=1.0\linewidth]{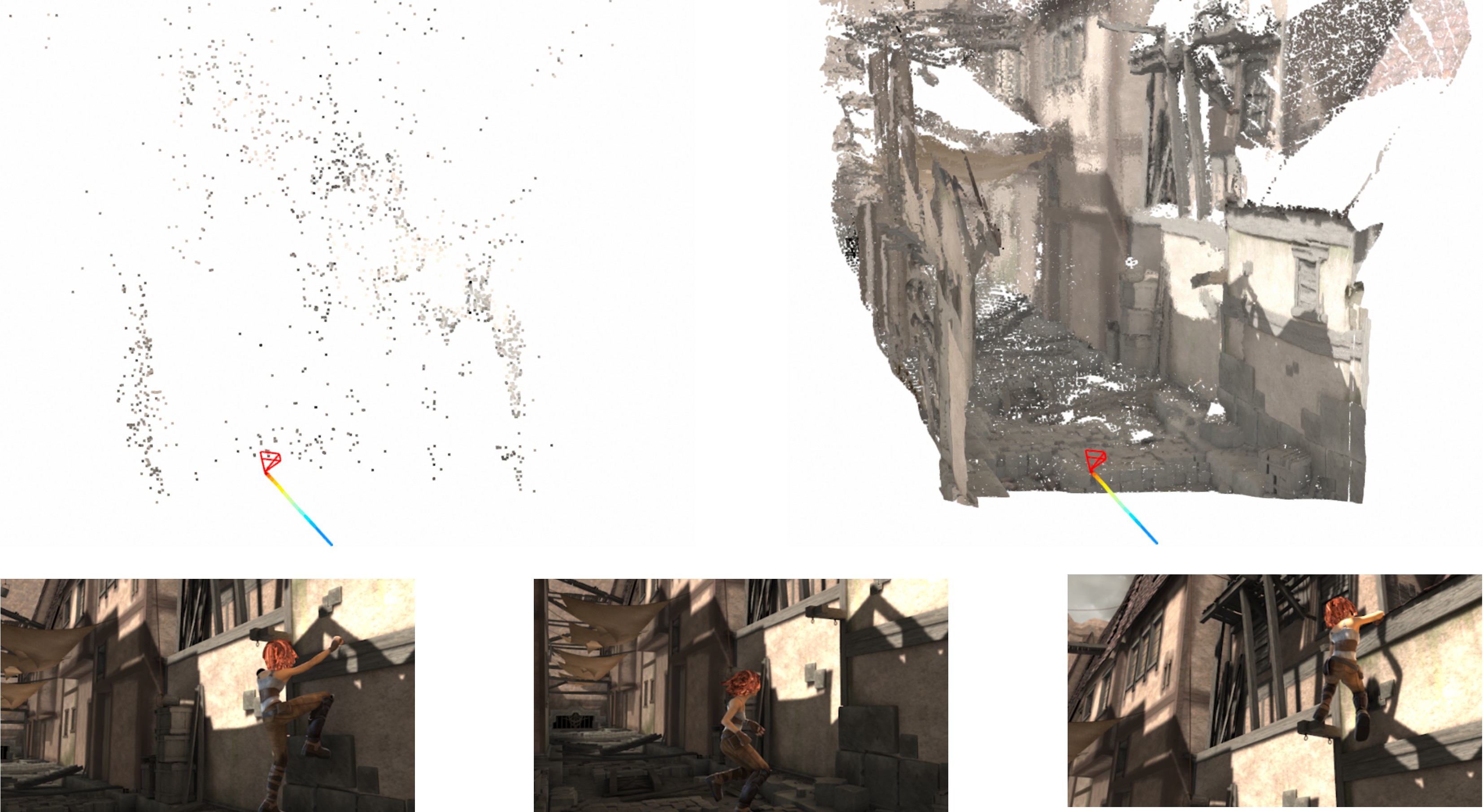}
      \caption{Qualitative results on the Sintel dataset~\cite{butler2012eccv}.}
  \end{subfigure}
  \\[0.5em]
  \begin{subfigure}[b]{\columnwidth}
      \centering
      \includegraphics[width=1.0\linewidth]{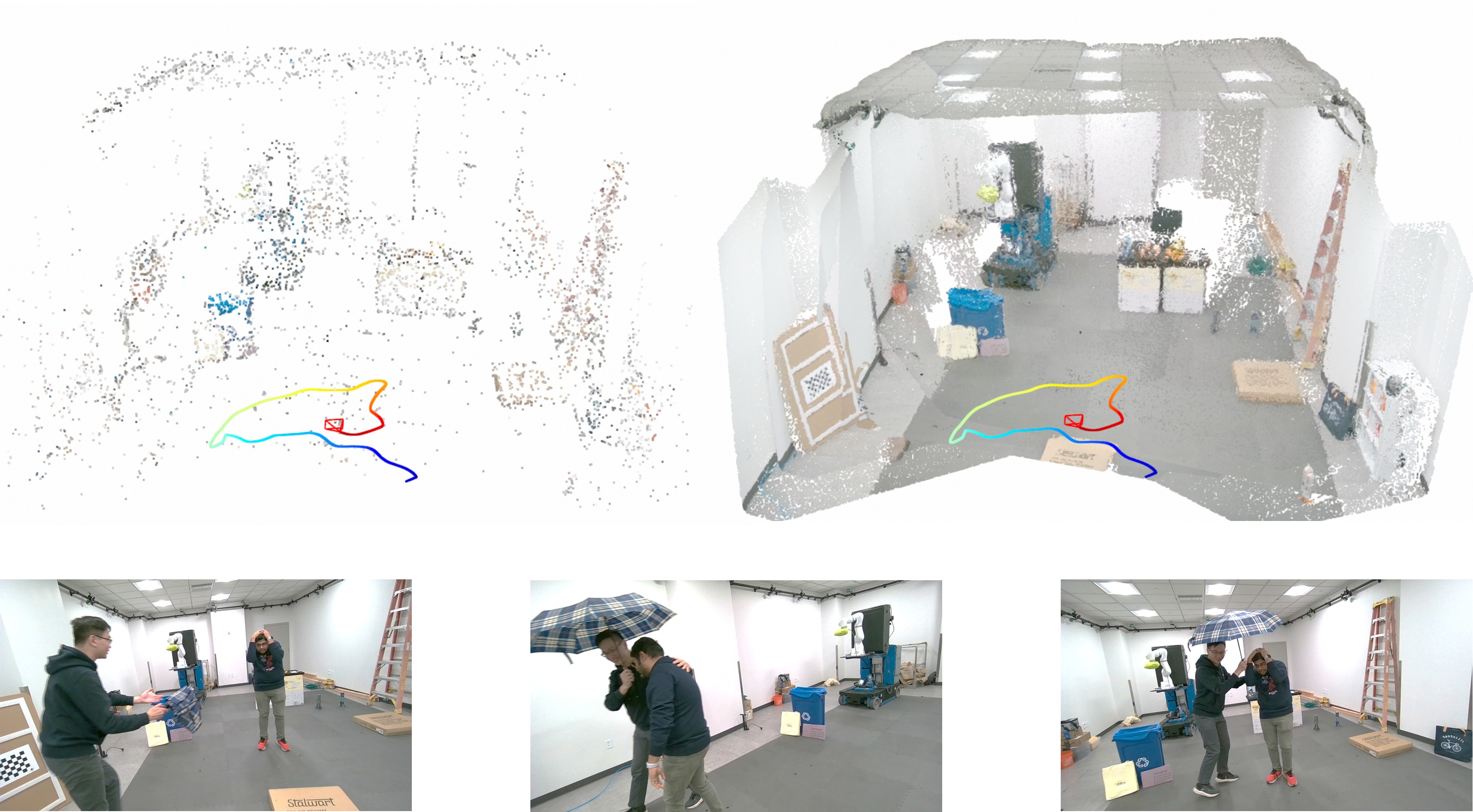}
      \caption{Qualitative results on the Wild-SLAM MoCap Dataset~\cite{palazzolo2019iros}.}
  \end{subfigure}
  \caption{Qualitative results on different datasets. We show the sparse patch point cloud map, the dense point cloud map accumulated from aligned depth predicted by $\pi^3_{\text{mos}}$, and the input data containing various dynamic objects.}
  \label{Fig:demo}
  \vspace{-0.3cm}
\end{figure}

\textbf{Results.} We present quantitative results in \cref{tab:tracking_performance_bonn}, \cref{tab:tracking_performance_wild} and \cref{tab:tracking_performance_sintel}. 
The results show that methods without explicit handling of moving objects~\cite{mur-artal2017tro, teed2021neurips, tartanvo2020corl} and purely geometry-based systems~\cite{palazzolo2019iros} struggle to run accurately in highly dynamic environments, yielding the lowest performance.
DynaSLAM~\cite{bescos2018ral} performs poorly when the moving objects exceed the detector's recognition capability, as observed in the \texttt{Table2} sequence of \cref{tab:tracking_performance_wild}.
MonST3R~\cite{zhang2025iclr} segments static backgrounds via optical flow and performs sliding-window registration, but due to prediction errors from both the feed-forward reconstruction and flow models, it exhibits significant drift over long sequences. 
Similarly, the offline reconstruction method MegaSaM~\cite{li2025cvpr-mafa} cannot process an entire sequence within the limited GPU memory (16 GB in our experimental setup), the necessary segmented alignment strategy results in unstable performance on longer sequences.
WildGS-SLAM~\cite{zheng2025cvpr-wsmg} performs comparably to ours in small-scale indoor scenes with limited camera motion range (as shown in \cref{tab:tracking_performance_bonn} and \cref{tab:tracking_performance_wild}).
But it relies on many overlapping views to identify distractors. Consequently, its performance degrades severely when the camera moves quickly or with little view overlap, as seen in the Sintel dataset (\cref{tab:tracking_performance_sintel}).
In contrast, our method achieves the best results across all three datasets, demonstrating superior robustness and accuracy under diverse dynamic conditions.
We show the system trajectories and reconstruction results on different datasets in \cref{Fig:demo}. It can be seen that our method robustly filters out dynamic objects and reconstructs clean maps.

\subsection{Video Depth Estimation}
\label{subsec:video depth}
\textbf{Datasets.} To evaluate the ability of our method for generating accurate and scale-consistent dense depth maps, we conducted video depth estimation experiments. Following the setup from CUT3R~\cite{wang2025cvpr-c3pm} and $\pi^3$~\cite{wang2025arxiv-pspe}, we used the Bonn \mbox{RGB-D} Dynamic Dataset~\cite{palazzolo2019iros} and the MPI Sintel Dataset~\cite{butler2012eccv} introduced in \cref{subsec:tracking}.
For the Bonn dataset, we selected five highly dynamic sequences, each containing 400--900 frames, while for Sintel we used the same 14 sequences as in \cref{subsec:tracking} consisting of 20--50 frames each.

\textbf{Metric and Baselines.} 
We use absolute relative error (Abs Rel) and percentage of predicted depths within a 1.25-factor of true depth ($\delta < 1.25$) as metrics. These metrics are evaluated under per-sequence scale alignment.
We mainly selected a diverse set of feed-forward reconstruction models as baselines. 
These include the offline models Fast3R~\cite{yang2025cvpr}, VGGT~\cite{wang2025cvpr-vvgg}, and $\pi^3$~\cite{wang2025arxiv-pspe}, which process entire sequences as a single batch. While this helps maintain better scale consistency, it comes at the cost of substantial GPU memory consumption.
In addition, we included several memory-efficient online models, such as pair-wise models DUSt3R~\cite{wang2024cvpr-dg3v}, MASt3R~\cite{leroy2024eccv}, and MonST3R~\cite{zhang2025iclr}, which fine-tuned on dynamic scenes.
We also evaluated incremental methods including Spann3R~\cite{wang2024threedv}, CUT3R~\cite{wang2025cvpr-c3pm}, Point3R~\cite{Wu2025arxiv-ps3r}, and StreamVGGT~\cite{zhuo2025arxiv-s4gt}, where Spann3R, CUT3R, and Point3R are extended from DUSt3R, and StreamVGGT is extended from VGGT. 

\textbf{Results.} As shown in \cref{tab:video_depth}, our method achieves the best performance among all online approaches on the Bonn dataset, with an Abs Rel error close to that of the best offline model, $\pi^3$.
On the Sintel dataset, our performance is slightly lower than $\pi^3$, yet our method still attains the lowest Abs Rel among all online baselines and ranks second in $\delta < 1.25$.
These results demonstrate that our approach can produce accurate, scale-consistent depth estimations and achieve performance comparable to state-of-the-art offline models.
% It is worth mentioning that although our system requires loading $\pi^3_{\text{mos}}$, the batch size for each inference is only 6 images, resulting in a modest memory footprint. In practice, the entire system runs stably on sequences exceeding 1000 frames within 11 GB of GPU memory.

\begin{table}[t]
  \centering
  \caption{Video depth estimation results on the Sintel~\cite{butler2012eccv} and Bonn~\cite{palazzolo2019iros} Datasets.}
  \label{tab:video_depth}
  \setlength{\tabcolsep}{4pt}
  \renewcommand{\arraystretch}{1.15}
  \resizebox{0.9\columnwidth}{!}{%
  \begin{tabular}{l|cc|cc}
  \toprule
  & \multicolumn{2}{c}{\texttt{Sintel}} & \multicolumn{2}{c}{\texttt{Bonn}} \\
  \cmidrule(lr){2-3}\cmidrule(lr){4-5}
  \textbf{Method} &
  Abs\ Rel $\downarrow$ & $\delta < 1.25 \uparrow$ &
  Abs\ Rel $\downarrow$ & $\delta < 1.25 \uparrow$\\
  \midrule
  \rowcolor{gray!15}
  \multicolumn{5}{l}{\textit{Offline Methods}} \\
  Fast3R~\cite{yang2025cvpr} & 0.638 & 0.422 & 0.194 & 0.772 \\
  VGGT~\cite{wang2025cvpr-vvgg} & \underline{0.299} & \underline{0.638} & \underline{0.057} & \underline{0.966} \\
  $\pi^3$~\cite{wang2025arxiv-pspe} & \textbf{0.233} & \textbf{0.664} & \textbf{0.049} & \textbf{0.975} \\
  \hline
  \rowcolor{gray!15}
  \multicolumn{5}{l}{\textit{Online Methods}} \\
  DUSt3R~\cite{wang2024cvpr-dg3v}      & 0.662 & 0.434 & 0.151 & 0.839 \\
  MASt3R~\cite{leroy2024eccv}    & 0.558 & 0.487 & 0.188 & 0.765 \\
  MonST3R~\cite{zhang2025iclr}        & 0.399 & 0.519 & 0.072 & 0.957 \\
  Spann3R~\cite{wang2024threedv}        & 0.622 & 0.426 & 0.144 & 0.813 \\
  CUT3R~\cite{wang2025cvpr-c3pm}        & 0.417 & 0.507 & 0.078 & 0.937 \\
  Point3R~\cite{Wu2025arxiv-ps3r}     & 0.452 & 0.489 & 0.060 & 0.960 \\
  StreamVGGT~\cite{zhuo2025arxiv-s4gt}  & \underline{0.323} & \textbf{0.657} & \underline{0.059} & \underline{0.972} \\
  \midrule
  \textbf{ours}        & \textbf{0.287} & \underline{0.625} & \textbf{0.054} & \textbf{0.985} \\
  \bottomrule
  \end{tabular}}
  \vspace{-0.2cm}
\end{table}

\subsection{Ablation Study}
\label{subsec:ablation}
We conduct ablation experiments of camera tracking performance on the Bonn \mbox{RGB-D} Dynamic Dataset~\cite{palazzolo2019iros} and the Wild-SLAM MoCap Dataset~\cite{palazzolo2019iros} to evaluate the impact of each component in our system. Specifically, we analyze three key modules introduced in \cref{sec:Method}: 
(1)~\textbf{Moving mask}, whether to use the moving object mask from $\pi^3_{\text{mos}}$ to exclude dynamic regions when selecting and tracking patches;
(2)~\textbf{Depth Prior}, whether to incorporate the scale-aligned depth map predicted by $\pi^3_{\text{mos}}$, which means use the depth prior loss $\mathcal{L}_{\text{p}}$ in \cref{subsec:weighting} or not.
(3)~\textbf{Uncertainty-aware BA}, whether to adaptively weight the depth prior loss $\mathcal{L}_{\text{p}}$ based on frame-wise uncertainty during bundle adjustment.
When the Depth Prior is enabled but Uncertainty-aware BA is disabled, we use a fixed loss weight of 1.0 for all frames.

It is worth noting that the patch-based SLAM framework underlying our system inherently filters out points with large reprojection errors to mitigate the influence of outliers.
Even so, as shown in \cref{tab:ablation}, disabling the moving object mask, as in settings (a) and (b), leads to a significant degradation in camera tracking accuracy. Moreover, incorporating the depth prior without masking may even introduce additional errors, since uncertain dynamic patches can remain in the optimization window, as reflected in the poor performance of (b) on the \texttt{Wild} dataset.
In contrast, using only the moving object mask, as in (c), already yields strong performance, highlighting the importance of the mask from $\pi^3_{\text{mos}}$ for robust operation in dynamic environments.
Adding a fixed-weight depth prior further improves accuracy slightly, while incorporating Uncertainty-aware BA achieves the best overall tracking performance.

\begin{table}[t]
  \caption{Ablation study of camera tracking performance on Bonn and Wild-SLAM datasets (ATE RMSE $\downarrow$ [cm]).}
  \centering
  \footnotesize
  \renewcommand{\arraystretch}{0.9}
  \resizebox{\linewidth}{!}{
  \begin{tabular}{cccc|cc} 
    \toprule
    {} & Moving mask & Depth prior & Uncertainty BA & \texttt{Bonn} & \texttt{Wild} \\  
    \midrule
    (a) & \ding{55} & \ding{55} & \ding{55} & 4.82 & 1.23  \\
    (b) & \ding{55} & \ding{51} & \ding{51} & 3.91 & 3.78 \\
    (c) & \ding{51} & \ding{55} & \ding{55} & 2.67	& 0.98 \\
    (d) & \ding{51} & \ding{51} & \ding{55} & \underline{2.52} &	\underline{0.82} \\
     \midrule
    (e)& \ding{51}& \ding{51} &\ding{51} & \textbf{2.20} &	\textbf{0.42}  \\
    \bottomrule  
  \end{tabular}
  }
  \label{tab:ablation}
    \vspace{-0.5cm}
\end{table}

\section{Conclusion}
\label{subsec:conclusion}
In this paper, we present a novel monocular visual SLAM system that can robustly estimate camera poses and predict scale-consistent depth maps in dynamic scenes. We proposed a feed-forward reconstruction model, called $\pi^3_{\text{mos}}$, to provide precise moving object masks, while also exploiting depth predictions to further robustify the patch-based SLAM. By aligning predicted depth with estimated patches, we robustly handle the inherent scale ambiguities of the feed-forward reconstruction model. Our work effectively combines the strengths of feed-forward reconstruction models and classical SLAM frameworks, enabling robust online operation in dynamic environments. 
{\small
\textbf{Limitations.}
Although our method achieves strong performance across multiple tasks, it requires running multi-frame feed-forward model inference for each frame. This limits the system to a frame rate of only 2~fps on an NVIDIA RTX5000, making real-time deployment in robotic applications challenging.}

% \textbf{Limitations.}

% \todo{Have full names of authors for references!}

{
  \small
  \bibliographystyle{ieeenat_fullname}
  \bibliography{new,glorified}
}

% WARNING: do not forget to delete the supplementary pages from your submission 
% \input{sec/X_suppl}

\IfFileExists{./certificate/certificate.tex}{
\subfile{./certificate/certificate.tex}
}{}
\end{document}